\documentclass[sn-mathphys,Numbered]{sn-jnl}% Math and Physical Sciences Reference Style
\usepackage{graphicx}%
\usepackage{multirow}%
\usepackage{amsmath,amssymb,amsfonts}%
\usepackage{amsthm}%
\usepackage{mathrsfs}%
\usepackage[title]{appendix}%
\usepackage{xcolor}%
\usepackage{textcomp}%
\usepackage{manyfoot}%
\usepackage{booktabs}%
\usepackage{algorithm}%
\usepackage{algorithmicx}%
\usepackage{algpseudocode}%
\usepackage{listings}%
\usepackage{changepage, bm, makecell, multirow,  setspace, color}
\usepackage{mathtools}
\usepackage{verbatim}
\usepackage{caption}
\usepackage{threeparttable}
\usepackage{booktabs} 
\usepackage{float} 
\usepackage{subcaption}
\usepackage{soul}
\usepackage{svg}
\theoremstyle{thmstyleone}%
\theoremstyle{thmstyletwo}%

\theoremstyle{thmstylethree}%

\begin{document}

% \title[room occupancy prediction]{room occupancy prediction}
\title{Room Occupancy Prediction: Exploring the Power of Machine Learning and Temporal Insights
}

%%=============================================================%%
%% Prefix	-> \pfx{Dr}
%% GivenName	-> \fnm{Joergen W.}
%% Particle	-> \spfx{van der} -> surname prefix
%% FamilyName	-> \sur{Ploeg}
%% Suffix	-> \sfx{IV}
%% NatureName	-> \tanm{Poet Laureate} -> Title after name
%% Degrees	-> \dgr{MSc, PhD}
%% \author*[1,2]{\pfx{Dr} \fnm{Joergen W.} \spfx{van der} \sur{Ploeg} \sfx{IV} \tanm{Poet Laureate} 
%%                 \dgr{MSc, PhD}}\email{iauthor@gmail.com}
%%=============================================================%%

% \author*[1,2]{\fnm{First} \sur{Author}}\email{iauthor@gmail.com}

% \author[2,3]{\fnm{Second} \sur{Author}}\email{iiauthor@gmail.com}
% \equalcont{These authors contributed equally to this work.}

% \author[1,2]{\fnm{Third} \sur{Author}}\email{iiiauthor@gmail.com}
% \equalcont{These authors contributed equally to this work.}

% \affil*[1]{\orgdiv{Department}, \orgname{Organization}, \orgaddress{\street{Street}, \city{City}, \postcode{100190}, \state{State}, \country{Country}}}

% \affil[2]{\orgdiv{Department}, \orgname{Organization}, \orgaddress{\street{Street}, \city{City}, \postcode{10587}, \state{State}, \country{Country}}}

% \affil[3]{\orgdiv{Department}, \orgname{Organization}, \orgaddress{\street{Street}, \city{City}, \postcode{610101}, \state{State}, \country{Country}}}

\author[1]{\fnm{Siqi Mao} }\email{mden17g@gmail.com}
\equalcont{These authors contributed equally to this work.}

\author[2]{\fnm{Yaping Yuan}}\email{yypyuan@gmail.com}
\equalcont{These authors contributed equally to this work.}

\author*[1]{\fnm{Yinpu Li}}\email{yinpulee@gmail.com}
\equalcont{These authors contributed equally to this work.}

\author[3]{{Ziren Wang} }\email{zaynw1013@gmail.com}

\author[1]{\fnm{Yuanxin Yao} }\email{yaoyuanxin2017@gmail.com}

\author[1]{\fnm{Yixin Kang}}\email{kangyixin512@gmail.com}

\affil[1]{\orgdiv{Department of Statistics}, \orgname{Florida State University}, \orgaddress{\street{600 W College Ave}, \city{Tallahassee}, \postcode{32306}, \state{FL}, \country{USA}}}

\affil[2]{\orgdiv{Department of Mathematics and Statistics},
\orgname{University of Massachusetts Amherst}, \orgaddress{\street{710 N. Pleasant Street}, \city{Amherst}, \postcode{01003}, \state{MA}, \country{USA}}}

\affil[3]{\orgdiv{Department of Statistics}, \orgname{Rice University}, \orgaddress{\street{6100 Main St}, \city{Houston}, \postcode{77005}, \state{TX}, \country{USA}}}

%%==================================%%
%% sample for unstructured abstract %%
%%==================================%%

\abstract{Energy conservation in buildings is a paramount concern to combat greenhouse gas emissions and combat climate change. The efficient management of room occupancy, involving actions like lighting control and climate adjustment, is a pivotal strategy to curtail energy consumption. In contexts where surveillance technology isn't viable, non-intrusive sensors are employed to estimate room occupancy. In this study, we present a predictive framework for room occupancy that leverages a diverse set of machine learning models, with Random Forest consistently achieving the highest predictive accuracy. Notably, this dataset encompasses both temporal and spatial dimensions, revealing a wealth of information. Intriguingly, our framework demonstrates robust performance even in the absence of explicit temporal modeling. These findings underscore the remarkable predictive power of traditional machine learning models. The success can be attributed to the presence of feature redundancy, the simplicity of linear spatial and temporal patterns, and the advantages of high-frequency data sampling. While these results are compelling, it's essential to remain open to the possibility that explicitly modeling the temporal dimension could unlock deeper insights or further enhance predictive capabilities in specific scenarios. In summary, our research not only validates the effectiveness of our prediction framework for continuous and classification tasks but also underscores the potential for improvements through the inclusion of temporal aspects. The study highlights the promise of machine learning in shaping energy-efficient practices and room occupancy management.}

\keywords{Room occupancy prediction, Classification, Support Vector Machine, Random Forest, XGBoost,  Internet-of-Things (IoT)}

%%\pacs[JEL Classification]{D8, H51}

%%\pacs[MSC Classification]{35A01, 65L10, 65L12, 65L20, 65L70}

\maketitle

\section{Introduction}
Buildings account for as much as $40\%$ of the world's total energy consumption \cite{cao2016building} and contribute to $30\%$ of greenhouse gas emissions \cite{huovila2009buildings}. Therefore, diminishing the energy consumption within the construction sector will significantly contribute to addressing global energy usage and environmental carbon emissions concerns \cite{jacobson2009review}. In the pursuit of creating energy-efficient and comfortable indoor environments, the management of Heating, Ventilation, and Air Conditioning (HVAC) systems has emerged as a pivotal area of research and application. As the world faces increasingly complex challenges related to climate change, resource conservation, and sustainability, the optimization of HVAC systems has become an imperative task for both researchers and practitioners. One of the fundamental aspects of HVAC system optimization is the intelligent prediction and adaptation of room occupancy patterns. Occupancy in a room or building directly influences the HVAC system's operation. Efficiently managing temperature, ventilation, and airflow in spaces with varying levels of occupancy can lead to substantial energy savings, reduce environmental impact, and enhance the overall quality of indoor environments. 

The advancement in Internet-of-Things (IoT) and communication technologies has played a pivotal role in facilitating the integration of various wireless technologies into occupancy sensing. These wireless technologies include Radio Frequency Identification (RFID) \cite{li2011performance}, Wi-Fi \cite{wang2018occupancy}, and Bluetooth Low Energy (BLE) \cite{tekler2019alternative}, all with the aim of enhancing the range, speed, precision, level of detail, and energy efficiency of existing sensing methods. Nevertheless, despite their merits, terminal-based approaches encounter constraints due to the necessity of deploying specialized sensors and implementing third-party software components, such as BLE beacons, Wi-Fi access points, and mobile applications. These requirements lead to increased implementation costs, and disruption of occupants' daily routines, and raise concerns regarding privacy.

Furthermore, there are other methods rely on passive sensing technologies, which encompass CO2 sensors \cite{nassif2012robust}, Passive Infrared (PIR) sensors \cite{raykov2016predicting}, ultrasonic detection sensors \cite{shih2015occupancy}, sound detection sensors \cite{uziel2013networked}, camera systems \cite{liu2013measuring}, and intelligent power meters \cite{tekler2022plug}. These non-terminal strategies indirectly accumulate occupancy data for specific zones within a building where these sensors are strategically positioned. 

Machine learning techniques have been widely used in numerous sectors recently and achieved gratifying results \cite{jordan2015machine}, \cite{li2022adaptive}, \cite{linero2022bayesian}, \cite{chen2023tests}, \cite{mao2022time}. In addition to the progress made in sensing technologies, the recent utilization of machine learning for occupancy detection has yielded significant enhancements in accuracy compared to conventional methods. These enhancements can be credited to machine learning algorithms' capacity to analyze extensive datasets gathered from the building and discern patterns and relationships with the building's occupancy. Kadouce et al. \cite{kadouche2010user} employed a support vector machine (SVM) classifier to deduce whether occupants were present or absent. They accomplished this by training the SVM with sensory datasets derived from a variety of sources, including motion sensors, pressure detectors, lighting controls, door and switch contactors, as well as flow meters. Sangogboye et al. \cite{sangogboye2016improving} focused on modeling occupant presence in two commercial buildings using machine learning techniques applied to motion data. Their findings demonstrated that SVM exhibited robust performance in accordance with their results. Razavi et al. \cite{razavi2019occupancy}, conducted an analysis of electricity consumption data from over 5000 residential homes. They assessed the performance of a diverse range of machine learning models, including Support Vector Machines (SVM), K-Nearest Neighbors (KNN), Random Forest (RF), Gradient Boosting (GB), and neural networks. Their objective was to predict both current and future occupancy status. In a separate investigation, Park et al. \cite{park2021lstm}  introduced an occupancy detection model based on Long Short-Term Memory (LSTM) architecture. They utilized energy consumption data obtained from smart plugs to discern the presence of occupants in residential homes.
Utilizing supervised learning techniques such as Random Forest, Decision Trees, and Bagging, Koklu et al. \cite{koklu2019tree} conducted occupancy detection and assessed the models' performance through classification accuracy metrics. Assessing the accuracy of classification involves the computation of true positives (correctly recognized examples belonging to the class), true negatives (correctly recognized examples not part of the class), false positives (incorrectly assigned examples to the class), and false negatives (examples not recognized as class examples).

The central objective of this investigation is to demonstrate the robust applicability of the comprehensive framework introduced in \cite{li2023beyond}. This framework aims to enlighten individuals with a strong background in data science, who may have limited exposure to the field of sensor research, and aid in their adaptation for practical use across various research domains. We have employed this prediction framework to address the challenge of room occupancy prediction. While we have updated the framework by incorporating various classification methods, we have adhered to the established methodology. Our current study provides empirical evidence of the framework's high performance in predictive tasks. Additionally, our research has unveiled a counter-intuitive revelation: machine learning models do not explicitly consider a temporal structure if the temporal structure of the target variable has been fully reflected in responses variable. This unexpected superiority challenges conventional assumptions and underscores the potential practical utility of machine learning models in the context of room occupancy prediction.

% Additionally, our research has unveiled a counter-intuitive revelation: machine learning models that do not explicitly consider temporal and spatial dependencies outperform spatial-temporal models. 

This research generalized a prediction framework which is proposed in \cite{li2023beyond}. An empirical data analysis is performed to test the framework by utilizing the dataset referenced in \cite{singh2018machine}. The dataset comprises more than 10,000 data points, each associated with 16 distinct features, representing the readings of specific sensors. The experiment incorporated five distinct sensor types: temperature, illumination, sound, $CO_2$, and passive infrared (PIR). The primary objective of this investigation is to accurately estimate the occupancy levels within a laboratory setting.

The remaining sections of the paper are organized as follows: In Section 2, we provide a comprehensive exploration of the acquired data and the prediction methodology. This section offers an extensive elucidation of the various models under the prediction framework in our study, including Multinomial Logistic Regression, Linear Discriminant Analysis, Multi-class Support Vector Machine (MSVM), Random Forest (RF), XGBoost, LightGBM, and a Multi-layer Perceptron classifier. It also encompasses a detailed description of the evaluation metrics we employed and presents a thorough overview of the entire implementation process. Furthermore, this section includes a comprehensive performance comparison across different models, followed by an in-depth analysis. Additionally, we incorporate a SHAP (SHapley Additive exPlanations) analysis within this section. Section 3 explores why time series models may not be necessary for this specific dataset from two perspectives. The first perspective addresses the redundancy of time-dependent information in the target variable, it already presents in responses variables of the dataset, while the second considers the sampling frequency.
Lastly, in Section 4, we summarize the key conclusions derived from this study and outline potential directions for future research.

\section{Data and Methodologies}\label{sec2}

\subsection{Data}

The data are collected based on the experiment from the research \cite{singh2018machine}. The purpose of the experiment was to estimate occupancy by the signal information from non-intrusive sensors. Seven sensor nodes and one edge node were arranged in the setup, and the sensor nodes used wireless transceivers to send data to the edge every thirty seconds. In this experiment, five distinct kinds of non-intrusive sensors were used: digital passive infrared (PIR), light, sound, temperature, and CO2. Calibration of the PIR, sound, and CO2 sensors required considerable labor. Before the CO2 sensor was used for the first time, zero-point calibration was done by hand. This involved keeping the sensor in a clean environment for more than 20 minutes and then lowering the calibration pin (HD pin) for more than 7 seconds. In essence, the sound sensor is a microphone with an analog amplifier of variable gain attached. Two trimpots are available for the PIR sensor: one for adjusting sensitivity and the other for adjusting how long the output remains high after motion detection. These two were both set to their maximum settings. The temperature, light, and sound sensors were located in sensor nodes S1–S4, a CO2 sensor was located in sensor node S5, and one PIR sensor each was placed on the ceiling ledges of S6 and S7 at an angle that optimized the sensor's field of view for motion detection. Over the course of four carefully planned days, the number of occupants in each room varied from zero to three. The manual recording of the room's occupancy count served as the basis for accuracy. The data description is provided in Table \ref{data_descr}. 
\begin{table}[h!]
\caption{Data Description}\label{data_descr}%
\begin{tabular}{@{}lccccc@{}}
\toprule
Variable Name &	Type& Description & Units \\
    \midrule

Date 	 &Date	&	 &	YYYY/MM/DD	\\
Time&		Date	&&		HH:MM:SS\\
S1\_Temp&		Continuous	&&		C\\
S2\_{Temp}	&Continuous		&&	C	\\
S3\_{Temp}	&Continuous		&&	C \\
S4\_{Temp}&		Continuous	&&		C	\\
S1\_{Light}&		Integer	&&		Lux \\
S2\_{Light}	&Integer		&&	Lux	\\
S3\_{Light}&		Integer		&&	Lux	\\
S4\_{Light}&	Integer	&&		Lux \\

S1\_{Sound} &	Continuous	&	amplifier output read by ADC&	Volts \\
S2\_{Sound}	&	Continuous	&	amplifier output read by ADC	&Volts \\
S3\_{Sound}	&	Continuous		&amplifier output read by ADC	&Volts\\
S4\_{Sound}	&	Continuous	&	amplifier output read by ADC&	Volts\\
S5\_{CO2}		&Integer		&	& PPM	\\
S5\_{CO2\_Slope}&		Continuous	&	Slope of CO2 values	&\\
S6\_{PIR}&		Binary  & Binary value conveying motion detection	&\\
S7\_{PIR}	&	Integer		&Binary value conveying motion detection	&\\
Room\_Occupancy\_Count	&Integer	&	Ground Truth	&  \\
\botrule
\end{tabular}
\footnotetext{Source: The table is summarized from the data source website \url{https://archive.ics.uci.edu/dataset/864/room+occupancy+estimation}.}
\end{table}

\subsection{Exploratory Data Analysis}

There are all numerical features and totally of 10129 entries. In this study, 30\% is selected as the test proportion. The summary statistics of the features are displayed in Table \ref{Summary_Statistics_1} and \ref{Summary_Statistics_2}. All features are numerical. In addition, there are no missing values identified in the whole dataset. The features ``Date" and ``Time" are not considered in the training set. The rationales and discussions are provided in the discussion sector \ref{Discussion}. The values of the target variable ``Room\_Occupancy\_Count" are integers and the range is from 0 to 3.

\begin{table}[h]
\caption{Summary Statistics for features (Training data; Part 1).}\label{Summary_Statistics_1}%
\begin{tabular}{@{}lrrrrrrrr@{}}

{Statistics} &  S1\_Temp &  S2\_Temp &  S3\_Temp &  S4\_Temp &  S1\_Light &  S2\_Light &  S3\_Light &  S4\_Light \\
\midrule
Count &  8103 &  8103 &  8103 &  8103 &   8103 &   8103 &   8103 &   8103 \\
Mean  &    25.45 &    25.55 &    25.05 &    25.75 &     24.92 &     25.39 &     33.97 &     13.09 \\
STD   &     0.35 &     0.59 &     0.43 &     0.36 &     50.52 &     66.47 &     58.00 &     19.43 \\
Min   &    24.94 &    24.75 &    24.44 &    24.94 &      0.00 &      0.00 &      0.00 &      0.00 \\
First Quartile  &    25.19 &    25.19 &    24.69 &    25.44 &      0.00 &      0.00 &      0.00 &      0.00 \\
Median   &    25.38 &    25.38 &    24.94 &    25.75 &      0.00 &      0.00 &      0.00 &      0.00 \\
Third Quartile   &    25.63 &    25.63 &    25.38 &    26.00 &     11.00 &     13.00 &     49.00 &     21.00 \\
Max   &    26.38 &    29.00 &    26.19 &    26.50 &    165.00 &    258.00 &    280.00 &     74.00 \\
\botrule
\end{tabular}
\end{table}

\begin{table}[h]
\caption{Summary Statistics for features (Training data; Part 2)}\label{Summary_Statistics_2}%
\begin{tabular}{@{}lrrrrrrrr@{}}
\toprule
{Statistics} &  S1\_Sound &  S2\_Sound &  S3\_Sound &  S4\_Sound &  S5\_CO2 &  S5\_CO2\_Slope &  S6\_PIR &  S7\_PIR \\
\midrule
Count &  8103 &  8103 &  8103 &  8103 &   8103 &   8103 &   8103 &   8103 \\
Mean  &      0.17 &      0.12 &      0.16 &      0.10 &  461.01 &         -0.02 &    0.09 &    0.08 \\
STD   &      0.32 &      0.26 &      0.41 &      0.12 &  201.13 &          1.17 &    0.28 &    0.27 \\
Min   &      0.06 &      0.04 &      0.04 &      0.05 &  345.00 &         -6.30 &    0.00 &    0.00 \\
First Quartile   &      0.07 &      0.05 &      0.06 &      0.06 &  355.00 &         -0.05 &    0.00 &    0.00 \\
Median   &      0.08 &      0.05 &      0.06 &      0.08 &  360.00 &          0.00 &    0.00 &    0.00 \\
Third Quartile   &      0.08 &      0.06 &      0.07 &      0.10 &  465.00 &          0.00 &    0.00 &    0.00 \\
Max   &      3.88 &      3.44 &      3.67 &      3.40 & 1270.00 &          8.98 &    1.00 &    1.00 \\
\bottomrule
\end{tabular}
\end{table}

\subsection{Methods and implementation}

The initial introduction of the prediction framework can be traced back to the study presented in \cite{li2023beyond}, where it was primarily employed for addressing a water quality prediction challenge, constituting a regression problem. In this paper, we extend and adapt the prediction framework to encompass classification problems, and its efficacy is evaluated using room occupancy prediction data.

% The prediction framework was first proposed in the research \cite{li2023beyond} to handle a water quality prediction problem (a regression problem). The prediction framework is modified and generalized to classification problems and tested by the room occupancy prediction data in this paper. 

\subsubsection{Candidate Methodologies}

\begin{description}
    
    \item[\textbf{Benchmark}]
    The benchmark model provides a basic performance threshold to compare against more complex models. It uniformly predicts the most frequent class in the training data for all samples. While simple, it establishes a minimum accuracy level to improve upon.

    \item[\textbf{Multinomial Logistic Regression.}] Multinomial logistic regression extends the logistic regression model to multi-class classification tasks. It estimates the probability of each class using a softmax activation function. The cross-entropy loss is optimized to find the regression coefficients. Regularization helps prevent overfitting.

    \item[\textbf{Linear Discriminant Analysis (LDA).}] LDA finds a linear combination of features that maximizes the separation between classes. It assumes normal distributions with equal covariance for each class. The within-class and between-class scatter matrices are computed. Eigendecomposition provides projection directions to transform the data before classification.

    \item[\textbf{Multi-layer Perceptron (MLP) Classifier.}] The MLP classifier uses a feedforward artificial neural network model. Multiple layers of nodes with nonlinear activation functions allow modeling complex relationships. The network is trained via backpropagation to optimize a cross-entropy loss function. Regularization methods and dropout prevent overfitting.

    \item[\textbf{Multiclass Support Vector Machine (MSVM).}] MSVM extends SVM for multiple classes using the one-vs-one or one-vs-all approaches. It maximizes the margin between classes by mapping inputs to a high-dimensional feature space with a kernel function. Optimization finds the maximum margin hyperplanes between classes.

    \item[\textbf{Random Forest (RF).}] RF trains an ensemble of decision trees on random subsets of features and data. Each tree makes an independent prediction and the results are aggregated through voting or averaging to make the overall prediction. Randomness reduces overfitting and improves generalization.

    \item[\textbf{LightGBM.}] LightGBM employs gradient boosting to train an ensemble of decision trees. Unlike traditional gradient boosting frameworks, trees are grown leaf-wise rather than level-wise to minimize loss and enhance accuracy. Through sampling, less informative instances are filtered out to mitigate noise. Additionally, multiple leaves can share the same child node to bolster generalization. Regularization is applied to curb overfitting, while parallel and GPU-based learning mechanisms significantly augment efficiency.

    \item[\textbf{XGBoost.}] XGBoost stands for eXtreme Gradient Boosting, and as the name suggests, it implements gradient boosted decision trees with a focus on speed and performance. A notable feature is the regularization of the objective function to simplify the model complexity. Employing quantile sketching, it approximates the tree learning process, thus accelerating training. The framework is designed for parallel and distributed computing, leveraging multiple CPUs and machines. Out-of-core computing is another feature, wherein disk space is utilized to enhance performance further. The algorithm is finely tuned for sparse data handling and allows extensive customization through hyperparameter tuning.

\end{description}

\subsection{Model Selection}
\subsubsection{Evaluation Metrics}
Since we approach this task as a classification problem, we employ Balanced Accuracy, F-1 Score, and Area Under ROC Curve (AUC) as our evaluation metrics. Given that it involves multiple classes, we additionally incorporate the sample size of each class as weights to consolidate the measurements into a single metric:

\begin{equation}
Balanced \ Accuracy =  \frac{Specificity + Recall}{2},
\end{equation}

\begin{equation}
F1-Score =  \frac{2 * Precision * Recall}{Precision + Recall},
\end{equation}

\begin{equation}
AUC  =  \frac{\sum_{i \in \{C = 1\} } rank_i - \frac{M(1+M)}{2}}{M \times N},
\end{equation}

\begin{equation}
Weighted \ Metric =  \frac{\sum_{j=1}^C w_j Metric_j}{\sum_{j=1}^C w_j},
\end{equation}

where $M$ is positive sample size and $N$ is negative sample size, and set $C$ represents different classes. In the context of all three metrics, our goal is to maximize them as much as possible. Balanced Accuracy takes into account imbalanced datasets, ensuring an accurate representation of model performance across different classes. F-1 Score provides a balance between precision and recall, making it particularly useful when dealing with situations where false positives and false negatives carry different costs. AUC, on the other hand, quantifies the model's ability to distinguish between positive and negative instances across various thresholds. These metrics collectively offer a comprehensive perspective on the effectiveness of classification models.

\subsection{Implementation}

This section outlines the proposed prediction methodology, detailing its development through hyperparameter optimization, prediction generation, and assessment using specific scoring metrics. The implementation process involves the following steps:

% By developing the models through hyperparameter tuning, making predictions, and evaluating their efficacy using specific scoring metrics, we describe the suggested prediction methodology in this section. 

\begin{enumerate}
    \item Selection of Scoring Function: To fine-tune and assess hyperparameters, we employ the 
    Area Under the Receiver Operating Characteristic Curve (AUC-ROC) for all candidate techniques.
    % Select the Scoring Function: For adjusting and evaluating the hyperparameters, we use the  Area Under the Receiver Operating Characteristic Curve for all candidate techniques.

    \item Defining Hyperparameter Space: Each method's hyperparameter space is defined by specifying the relevant parameters.
    % Create the hyperparameter space for each method by defining the parameters.

    \item Cross-validation: We utilize cross-validation to perform hyperparameter optimization, aiming to identify the optimal hyperparameters for each method.
% Cross-validation is used to perform hyperparameter tweaking in order to identify the ideal hyperparameters for each method.

    \item Performance Evaluation: The evaluation metrics are computed on the test dataset for the best-tuned models within each method.

% Performance evaluation: Evaluate the evaluation metrics on the test data for each method of the best-tuned models.

    \item Results Storage: The outcomes, including the performance of each approach and the significance of its features, are saved for facilitating model comparison.
    
    % Save Results: Save the results of the effectiveness of each approach and the significance of its features for model comparison.
\end{enumerate}

\begin{figure}[hbt!]
  \centering
  \includegraphics[scale=1.0]{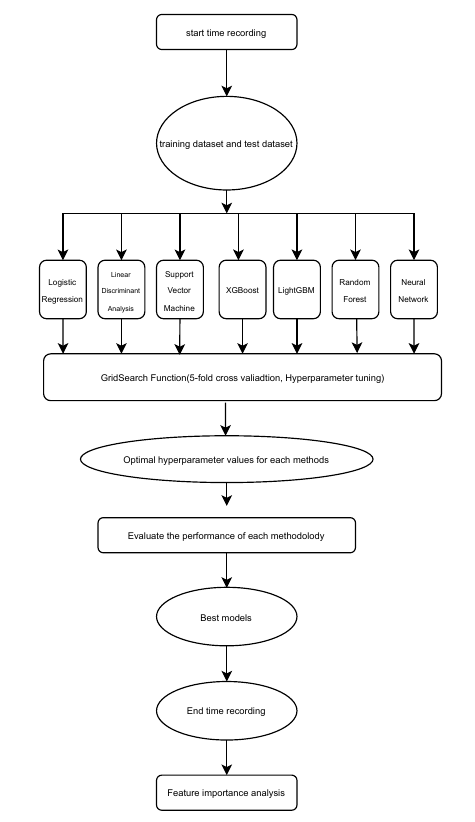}
  \caption{First, we partition the initial training dataset into training and testing subsets. Then, utilizing the training data, we perform hyperparameter tuning via GridSearchCV to identify the optimal parameters and models. We utilize these best models to make predictions on the original test dataset. Following the prediction phase, we proceed to select the best models for conducting feature importance and interaction analysis, leveraging the SHAP (SHapley Additive exPlanations) method.}
\end{figure}

\subsubsection{Results}

The prediction results are summarized in this section. The $5$-fold cross validation is selected in the hyperparameter tuning step.  Due to the sensitivity of machine learning algorithms to the scale of the features, standardization is applied to all the features. 

The model performance summary is displayed in Table \ref{prediction_numerical_only_std}. The model is sorted by the Weighted AUC score. Based on the performance of Benchmark model, we can see the necessary of employing advanced models is essential. Furthermore, the Random forest achieves the best performance under the selected metrics.

The hyper-parameters tuning time is presented in Table \ref{Time_prediction_all}. The number of fits in each model's hyper-parameter tuning is indicated by the term ``Total Fits" in the tables. The term ``Tuning Time" refers to the duration of each model's whole hyper-parameter tuning operation. ``Fitting Time (Best Model)" refers to the amount of time spent training each model's best hyper-parameters. The term ``Average Tuning" refers to the typical tuning time for hyper-parameters. From the the tuning time table, the tree methods have satisfying results and efficiency.

\begin{table}[h]
\caption{Room Occupancy Prediction Results (Numerical Features: Standardized).}\label{prediction_numerical_only_std}%
\begin{tabular}{@{}lccc@{}}
\toprule
{Models} &  Weighted $F_1$-Score &   Weighted AUC &  Balanced Accuracy \\
\midrule
Benchmark &  709.88	& N/A & 250.00\\
Logistic Regression            &       981.77 &        996.54 &                   941.55  \\
Linear Discriminant Analysis      &       981.59 &        997.14 &                   942.16 \\
Support Vector Machine            &       992.14 &        999.51 &                   976.35  \\
Multi-layer Perceptron classifier &       997.04 &        999.88 &                   991.96  \\
LightGBM                          &       997.03 &        999.94 &                   990.07  \\
XGBoost                           &       997.03 &        999.94 &                   989.81  \\
Random forest                     &       \textcolor{red}{998.52} &        \textcolor{red}{999.96} &                   \textcolor{red}{994.90} \\ 
\bottomrule
\end{tabular}

\footnotetext{Note. The numbers in \textcolor{red}{red} color stand for the best performance under the specific metric. All values are their original values $\times$ 1000.}

\end{table}

\begin{table}[h]
\caption{Running Time Summary in Seconds (Numerical Features: Standardized).}\label{Time_prediction_all}%
\begin{tabular}{@{}lcccc@{}}
\toprule
{Models} &  Total Fits &  Tuning Time&  Average Tuning  &  Fitting Time (Best Model) \\
    \midrule
Linear Discriminant Analysis      &              15 &         0.15 &                     0.03 &            0.01 \\
Random forest                     &            1200 &       120.17 &                     0.53 &            0.10 \\
Logistic Regression               &              15 &         2.04 &                     0.10 &            0.14 \\
LightGBM                          &            2000 &       295.56 &                     0.23 &            0.15 \\
Support Bector Machine            &              20 &         3.27 &                     1.36 &            0.16 \\
XGBoost                           &            5760 &      3199.18 &                     4.52 &            0.56 \\
Multi-layer Perceptron classifier &              40 &      4007.15 &                    97.81 &          100.18 \\
\bottomrule
\end{tabular}
\footnotetext{Note. The unit for ruining time is second.}
\end{table}

\subsection{SHapley Additive exPlanations (SHAP)}
For single predicted point, The localized contributions to that specific forecast are quantified by the SHAP values \cite{NIPS2017_8a20a862}. We provide the definition for the SHAP value:

\begin{align*}
\Phi_i(F, x) &= \sum_{Z'\subseteq X'} [ |Z'|\times \left(\begin{array}{l}M \\ |Z'|\end{array}\right)]^{-1}[F(Z')-F(Z'\backslash x_i)] \\
&=\sum_{Z'\subseteq X'}\frac{(|Z'|-1)!(M-|Z'|)!}{M!}[F(Z')-F(Z'\backslash x_i)].
\end{align*}

In this equation, $\Phi_i(F, x)$ represents the SHAP value corresponding to feature $x_i$ within the context of a model $F$ constructed using a set of features $X$. Here, $M$ stands for the total number of input features, $X'$ denotes the set of all potential feature combinations that include feature $x_i$, and $|Z'|$ represents the number of features within a specific feature combination denoted as $Z'$. Furthermore, $F(Z')$ and $F(Z'\backslash x_i)$ correspond to distinct predictive models trained on $|Z'|$ and $Z'$ with feature $x_i$ removed, respectively. Therefore, the SHAP value is determined by aggregating the marginal contributions $F(Z') - F(Z'\backslash x_i)$ from all feasible feature combinations $(Z')$ through a weighted average.

SHAP values are calculated using the Python implementation of SHAP, as described by Lundberg and Lee in \cite{NIPS2017_8a20a862}. In this study, we computed SHAP values for several candidate models.

\begin{figure}[hbt!]		\centering
        \includegraphics[width=0.6\linewidth]{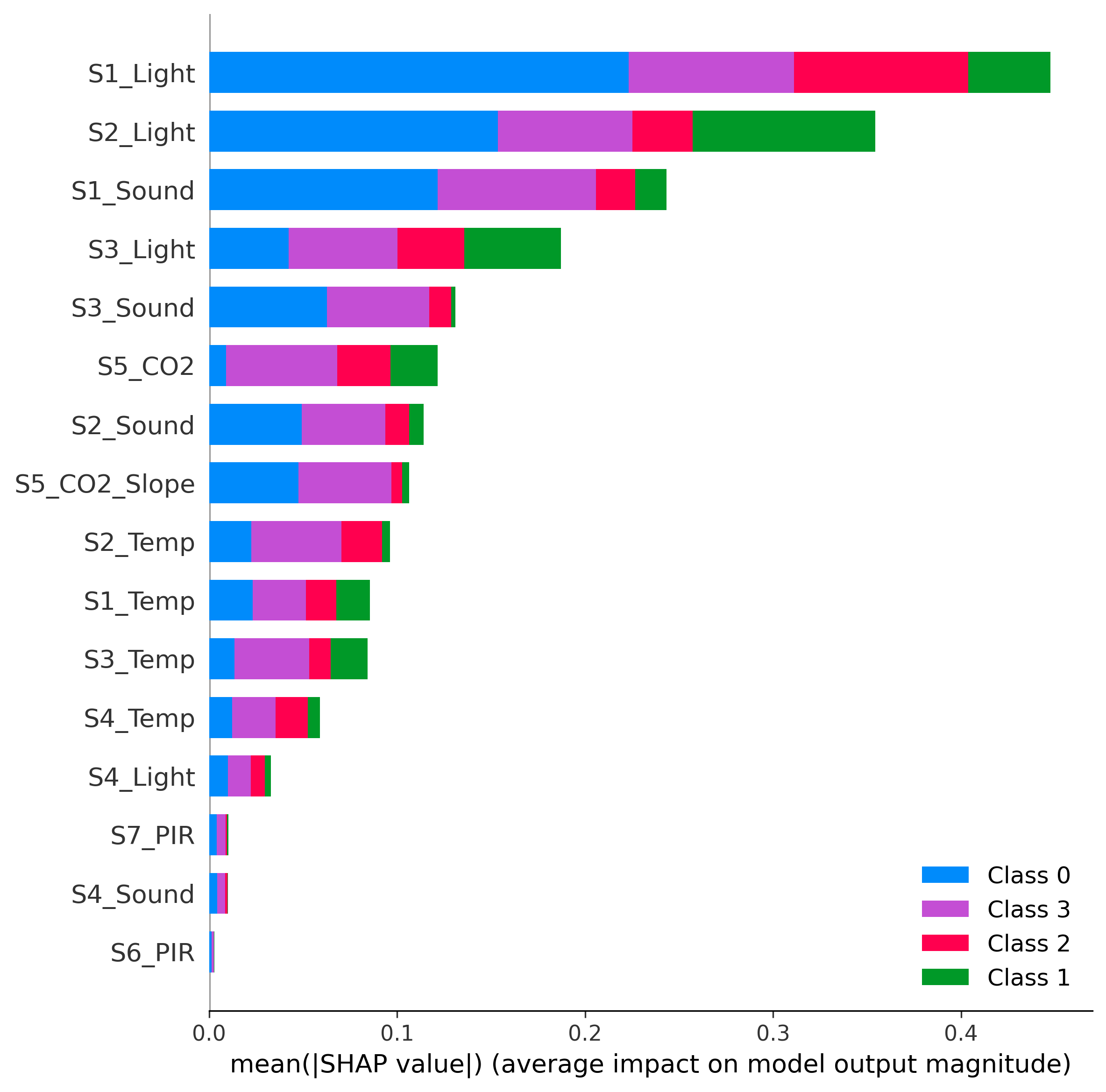}
    \caption{Average Impact (SHAP value) on model output for Random Forest.}\label{best_rf_shap_bar}
\end{figure}

Figure \ref{best_rf_shap_bar} is the illustration of SHAP values for the random forest model (best model). From the plot, the features ``S1\_Light" and ``S2\_Light" have notable impact in forecasting the room occupancy. Their impact are much higher than the other features.

\section{Discussion, Conclusion and Future Work}
\label{Discussion}

\subsection{Further Data Information Discussion: Redundancy in Temporal Features}

% \subsection{Redundancy Time Dependency Information}

Throughout our study, the dataset has exhibited both temporal and spatial dimensions. Surprisingly, our prediction framework has demonstrated strong performance without explicitly incorporating the temporal dimension. 

% We present the following arguments to shed light on this intriguing observation.

\begin{figure}[hbt!]
    \centering
    \includegraphics[width=0.9\textwidth]{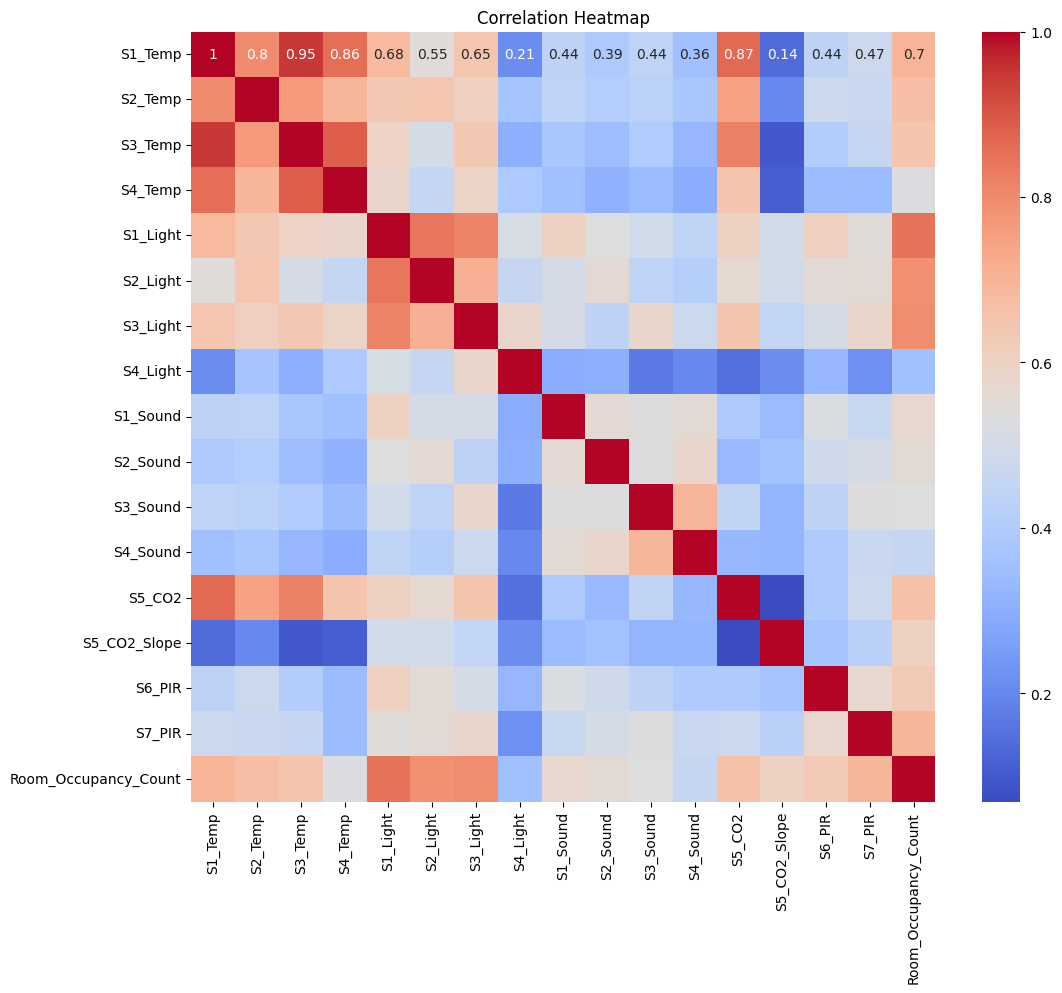}
    \caption{Correlation Coefficients Heatmap of Features.}
    \label{fig:corr}
\end{figure}

Our analysis reveals intriguing insights into the temporal aspects of our dataset, shedding light on the potential redundancy and high correlation (illustrated in Figure \ref{fig:corr}) between certain features. Notably, the autocorrelation results (detailed in Table \ref{tab1}) of our numeric columns provide evidence of the underlying temporal dynamics. For instance, let's consider the `S5\_CO2' column, which pertains to carbon dioxide levels. It exhibits an extraordinarily high autocorrelation coefficient of approximately 0.999. This remarkable finding underscores that the current $CO_2$ level is intimately connected with its past values. Essentially, as more individuals enter a room, the `S5\_CO2' readings tend to follow a consistent pattern of escalation, while their departure leads to a noticeable decline in the readings. Similarly, the `S5\_CO2\_Slope' representing the rate of change in $CO_2$ levels, also demonstrates a strong positive autocorrelation with a coefficient of about 0.998. This implies that the temporal aspect of our data, encapsulated by the $CO_2$ slope, is inherently linked to its past values. As more individuals arrive, the $CO_2$ slope tends to exhibit a higher positive value, whereas their departure results in a more pronounced negative slope. Furthermore, it's intriguing to note that various other sensors, such as temperature and light sensors (`S1\_Temp,' `S2\_Temp,' `S3\_Light,' etc.), display substantial autocorrelation coefficients, indicating consistent temporal patterns in their readings over time. This pattern of high autocorrelation coefficients across multiple sensors suggests that our dataset may indeed have inherent redundancy or temporal dependencies. 

% It's important to highlight that traditional machine learning models have demonstrated their capability to discern these underlying temporal patterns, even when the specific significance of each feature is not explicitly defined.

In essence, our findings support the notion that the temporal and spatial cues within the dataset might be subtly embedded within other features. The machine learning model can adeptly learn and leverage these temporal dynamics to make accurate predictions, emphasizing the richness of information present in our data.

Furthermore, our dataset exhibits a notably high sampling frequency in the temporal dimension, with data points recorded at 30-second intervals. This high-frequency data collection could potentially empower traditional models to implicitly incorporate temporal and spatial information, capitalizing on the fine granularity of the data points.

\begin{table}[h]
\caption{Autocorrelation Coefficients of the non-time Features}\label{tab1}%
\begin{tabular}{@{}ll@{}}
\toprule
Autocorrelation         & Results:  \\
\midrule

S1\_Temp:               & 0.9968  \\
S2\_Temp:               & 0.9982  \\
S3\_Temp:               & 0.9975  \\
S4\_Temp:               & 0.9939  \\
S1\_Light:              & 0.9975  \\
S2\_Light:              & 0.9955  \\
S3\_Light:              & 0.9970  \\
S4\_Light:              & 0.9996 \\
S1\_Sound:              & 0.4947 \\
S2\_Sound:              & 0.5516  \\
S3\_Sound:              & 0.5672  \\
S4\_Sound:              & 0.4689   \\
S5\_CO2:                & 0.9999 \\
S5\_CO2\_Slope:         & 0.9978 \\
S6\_PIR:                & 0.6015   \\
S7\_PIR:                & 0.7439 \\
Room\_Occupancy\_Count: & 0.9962 \\
\botrule
\end{tabular}
\end{table}

\begin{figure}[hbt!]
    \centering
    \begin{subfigure}{0.25\textwidth} % Adjust the width as needed
        \includegraphics[width=\linewidth]{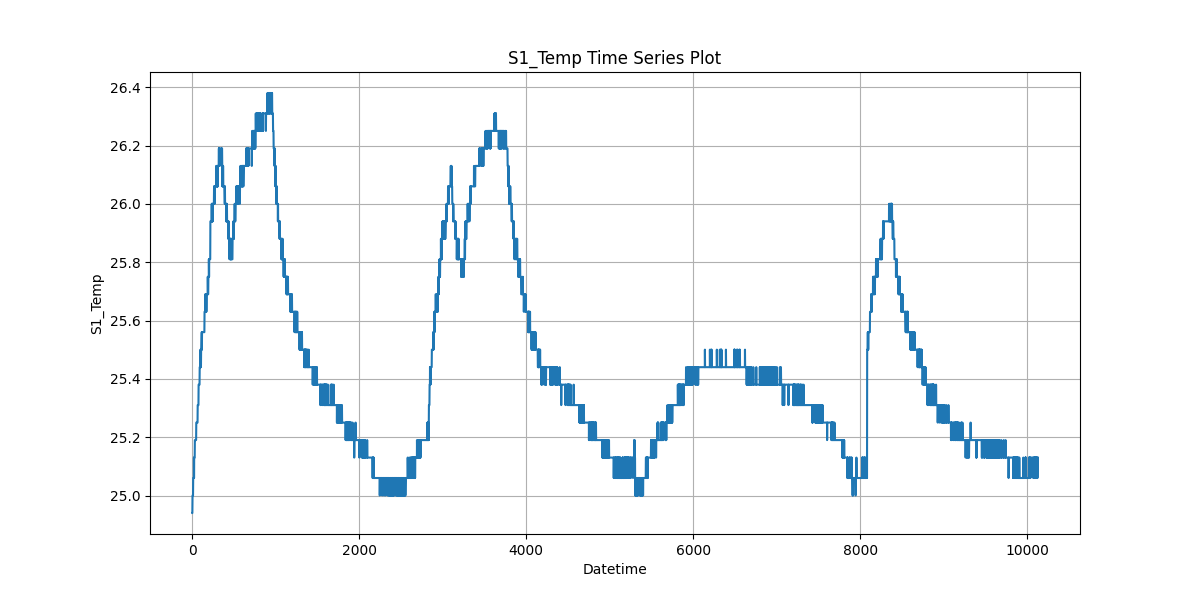}
        \caption{S1\_Temp}
        \label{fig:s1_temp}
    \end{subfigure}%
    \begin{subfigure}{0.25\textwidth} % Adjust the width as needed
        \includegraphics[width=\linewidth]{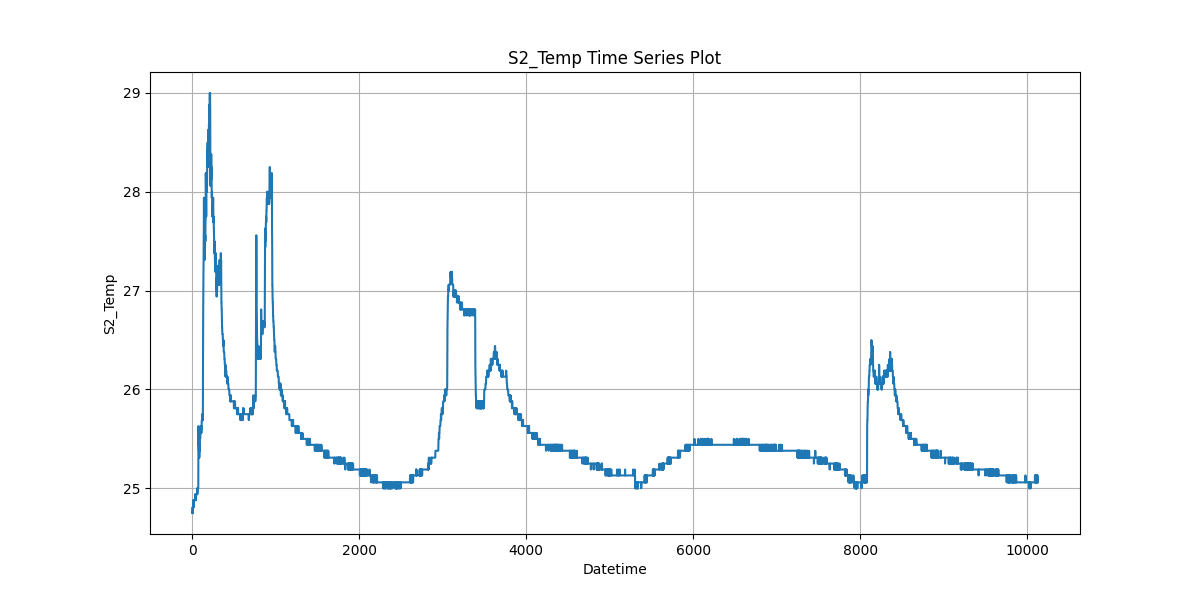}
        \caption{S2\_Temp}
    \end{subfigure}%
    \begin{subfigure}{0.25\textwidth} % Adjust the width as needed
        \includegraphics[width=\linewidth]{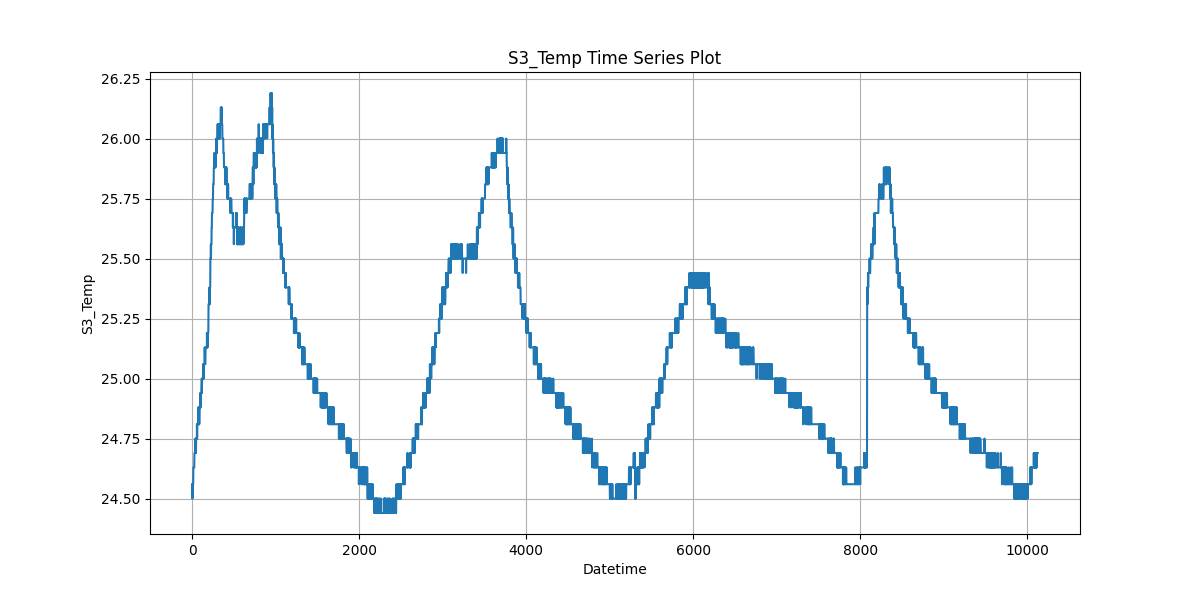}
        \caption{S3\_Temp}
    \end{subfigure}%
    \begin{subfigure}{0.25\textwidth} % Adjust the width as needed
        \includegraphics[width=\linewidth]{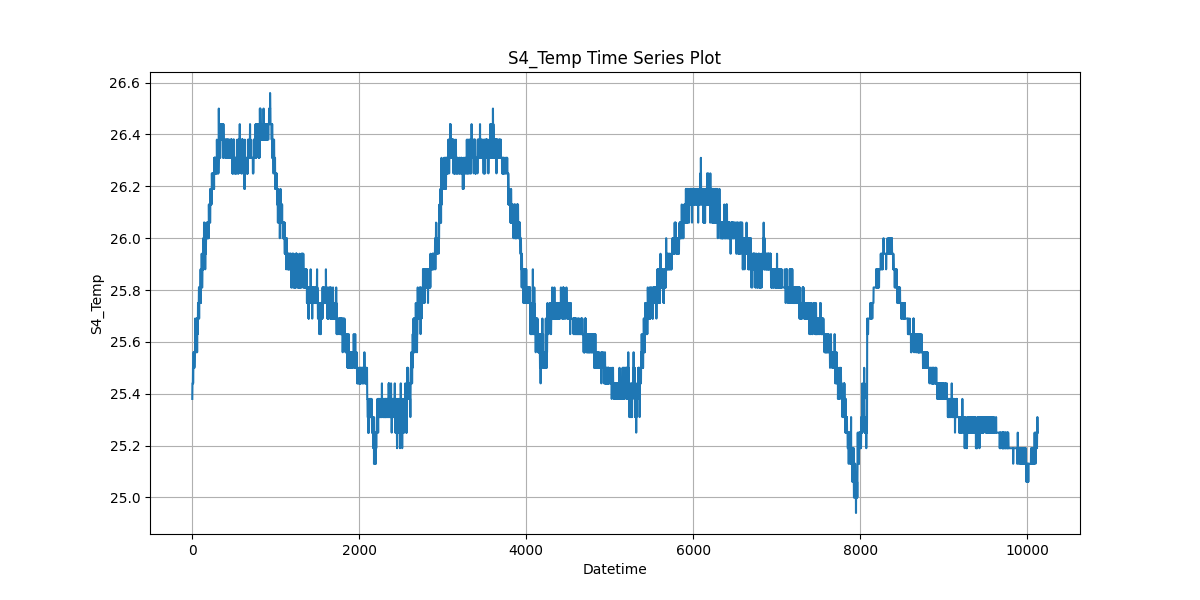}
        \caption{S4\_Temp}
    \end{subfigure}%
    \newline
    \begin{subfigure}{0.25\textwidth} % Adjust the width as needed
        \includegraphics[width=\linewidth]{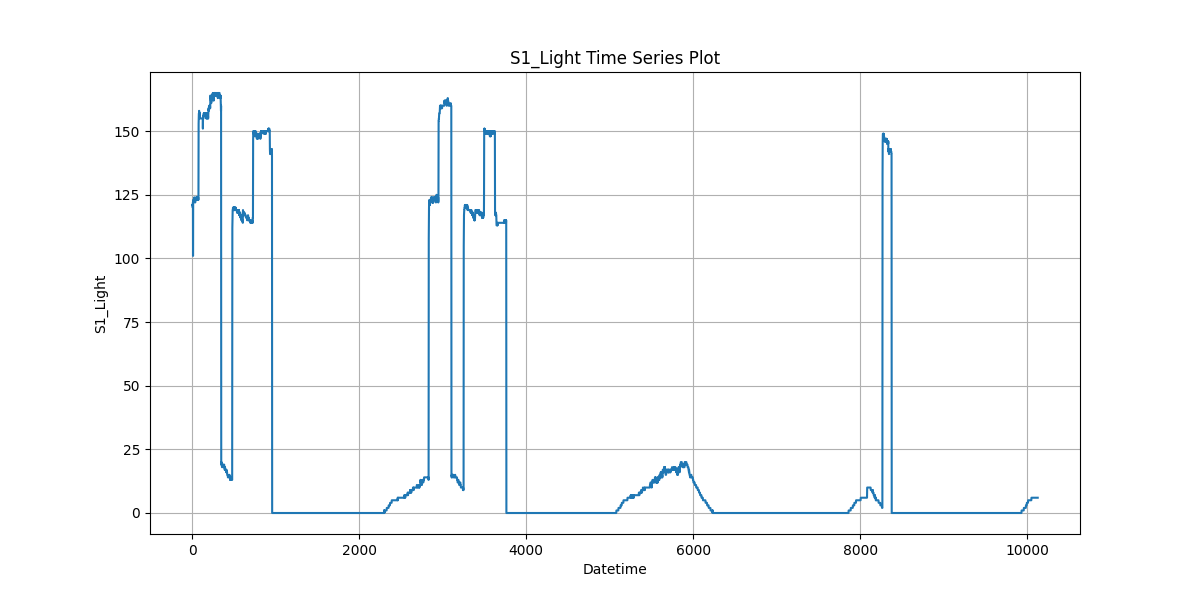}
        \caption{S1\_Light}
    \end{subfigure}%
    \begin{subfigure}{0.25\textwidth} % Adjust the width as needed
        \includegraphics[width=\linewidth]{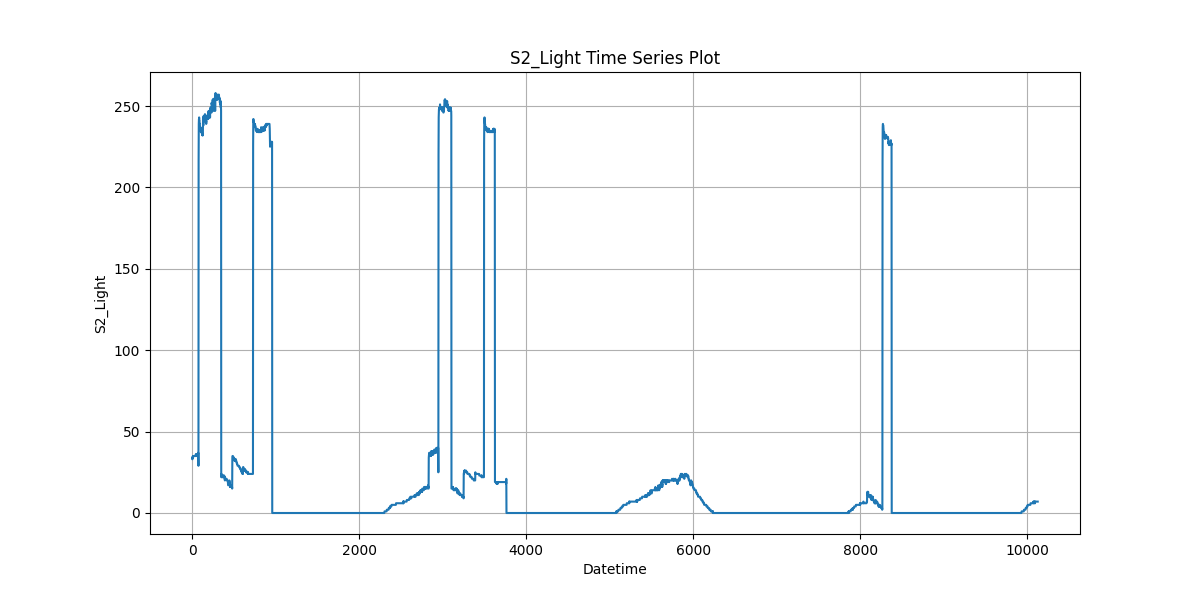}
        \caption{S2\_Light}
    \end{subfigure}%
    \begin{subfigure}{0.25\textwidth} % Adjust the width as needed
        \includegraphics[width=\linewidth]{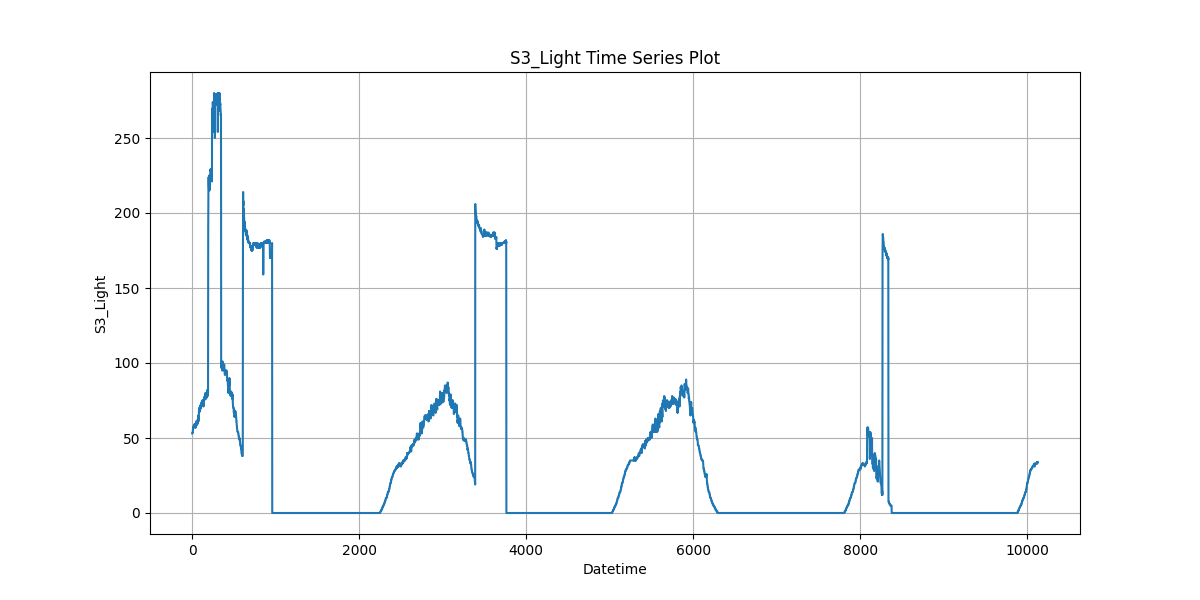}
        \caption{S3\_Light}
    \end{subfigure}%
    \begin{subfigure}{0.25\textwidth} % Adjust the width as needed
        \includegraphics[width=\linewidth]{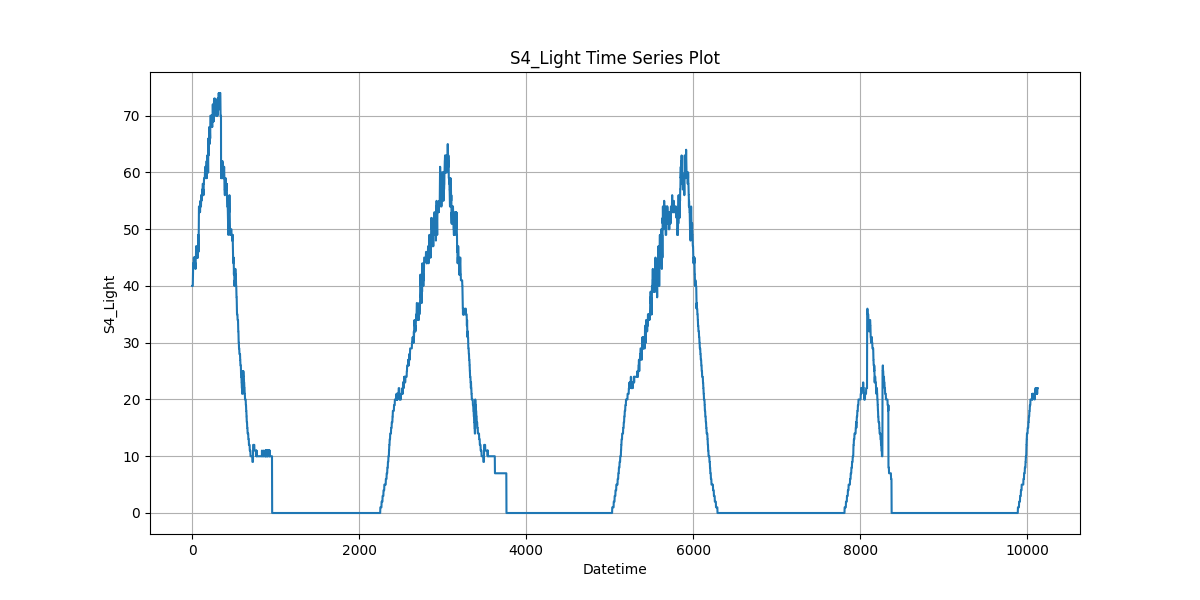}
        \caption{S4\_Light}
        \label{fig:s4_light}
    \end{subfigure}%
    \newline
    \begin{subfigure}{0.25\textwidth} % Adjust the width as needed
        \includegraphics[width=\linewidth]{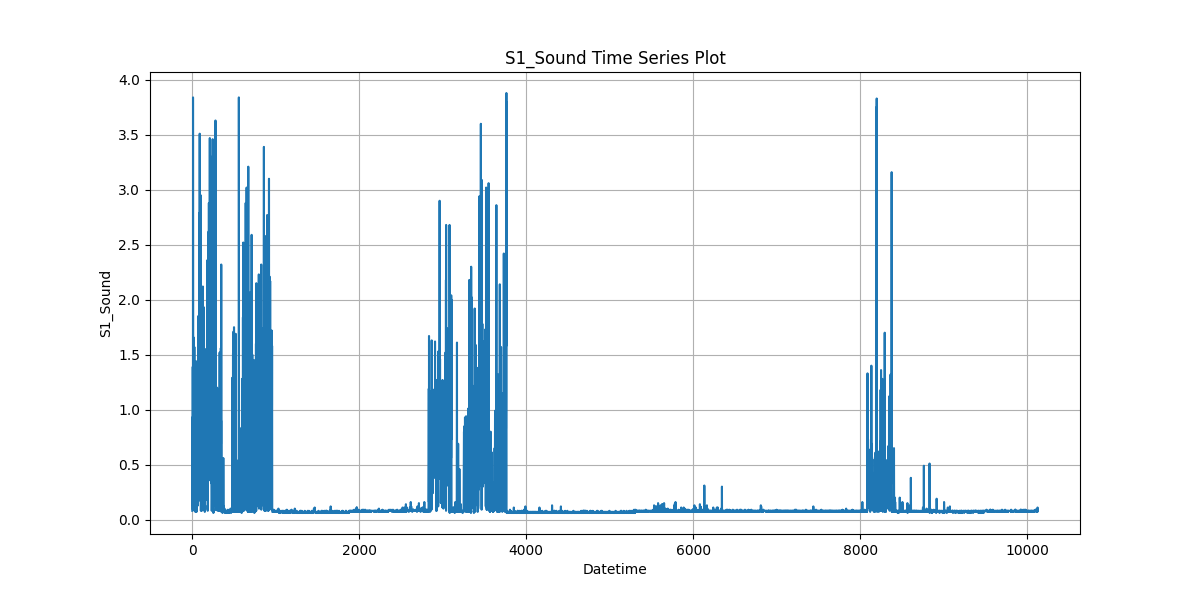}
        \caption{S1\_Sound}
        \label{fig:s1_sound}
    \end{subfigure}%
    \begin{subfigure}{0.25\textwidth} % Adjust the width as needed
        \includegraphics[width=\linewidth]{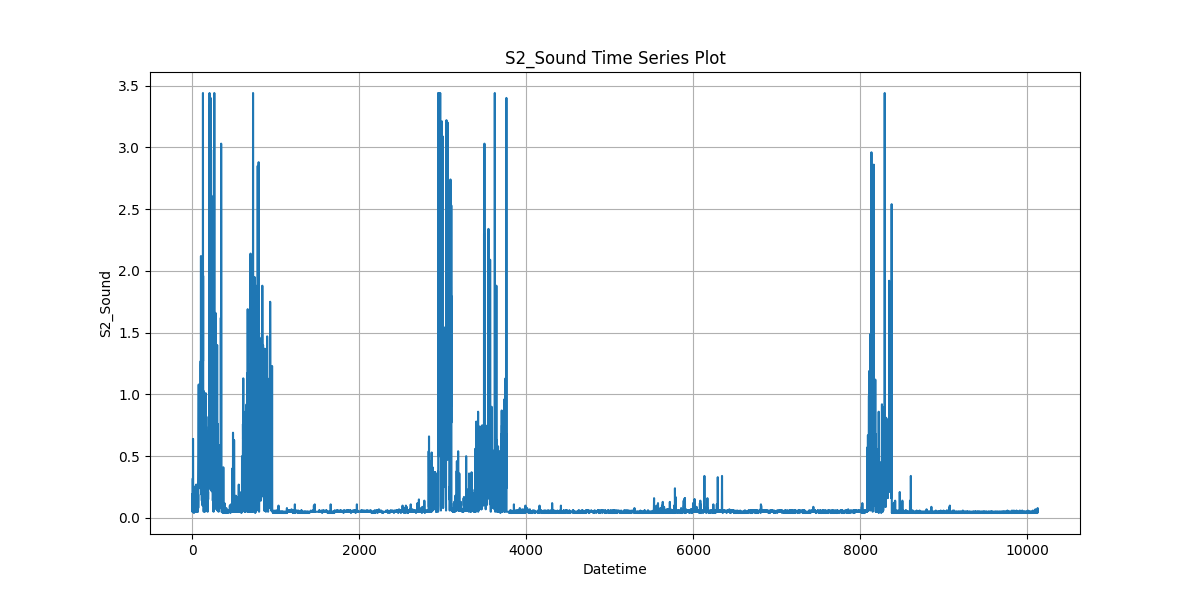}
        \caption{S2\_Sound}
        \label{fig:s2_sound}
    \end{subfigure}%
    \begin{subfigure}{0.25\textwidth} % Adjust the width as needed
        \includegraphics[width=\linewidth]{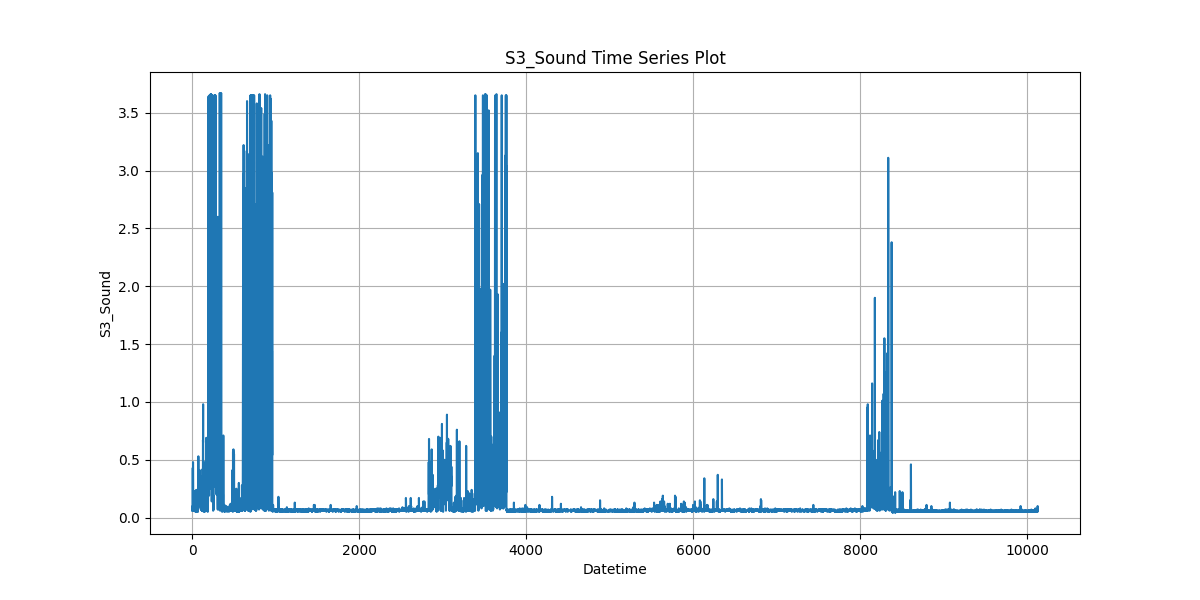}
        \caption{S3\_Sound}
        \label{fig:s3_sound}
    \end{subfigure}%
    \begin{subfigure}{0.25\textwidth} % Adjust the width as needed
        \includegraphics[width=\linewidth]{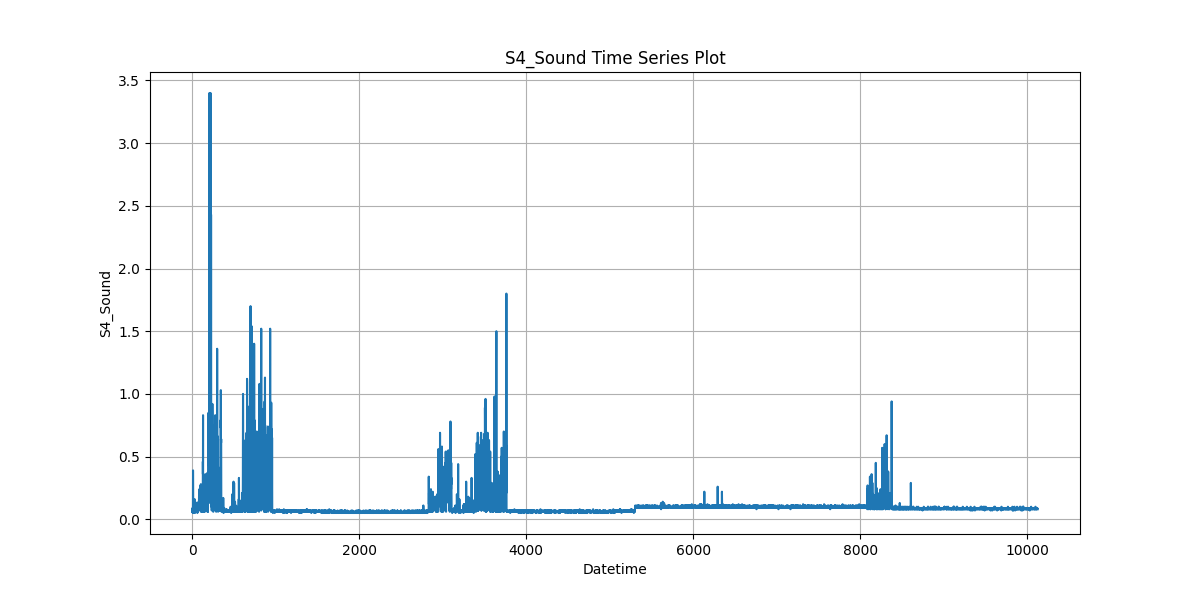}
        \caption{S4\_Sound}
        \label{fig:s4_sound}
    \end{subfigure}

    \caption{Time Series Plots for Selected Sensors.}
    \label{fig:sensor_time_series}
\end{figure}

The time series plots in Figure \ref{fig:sensor_time_series} illustrate the temporal patterns of various sensor data columns. These patterns are crucial in understanding the concept of redundant time information, where certain features exhibit high autocorrelation, indicating a strong linear relationship with their past values.

For instance, consider Figure \ref{fig:s1_temp}, which shows the time series plot for `S1\_Temp,' representing the temperature at Sensor 1. The consistently high autocorrelation observed in this plot suggests a significant temporal dependency, implying that the current temperature is closely related to its past values. Similar patterns can be observed in other sensors, such as `S2\_Temp,' `S3\_Temp,' and `S4\_Temp.'

These findings emphasize that the dataset contains valuable temporal information that traditional machine learning models can leverage, even when the specific significance of each feature is not explicitly defined. This temporal redundancy plays a vital role in understanding and predicting the behavior of our data.

% \item Linear Spatial and Temporal Patterns:
    
 Another noteworthy observation is that the spatial and temporal patterns within the dataset may follow relatively straightforward and linear relationships. In such cases, traditional machine learning models like logistic regression or decision trees might suffice in capturing these patterns without necessitating specialized spatial or temporal modeling. However, without comprehensive knowledge of the data collection, recording, and sensor aggregation procedures, we may not directly discern these patterns.

\begin{figure}[hbt!]%
\centering
\includegraphics[width=0.9\textwidth]{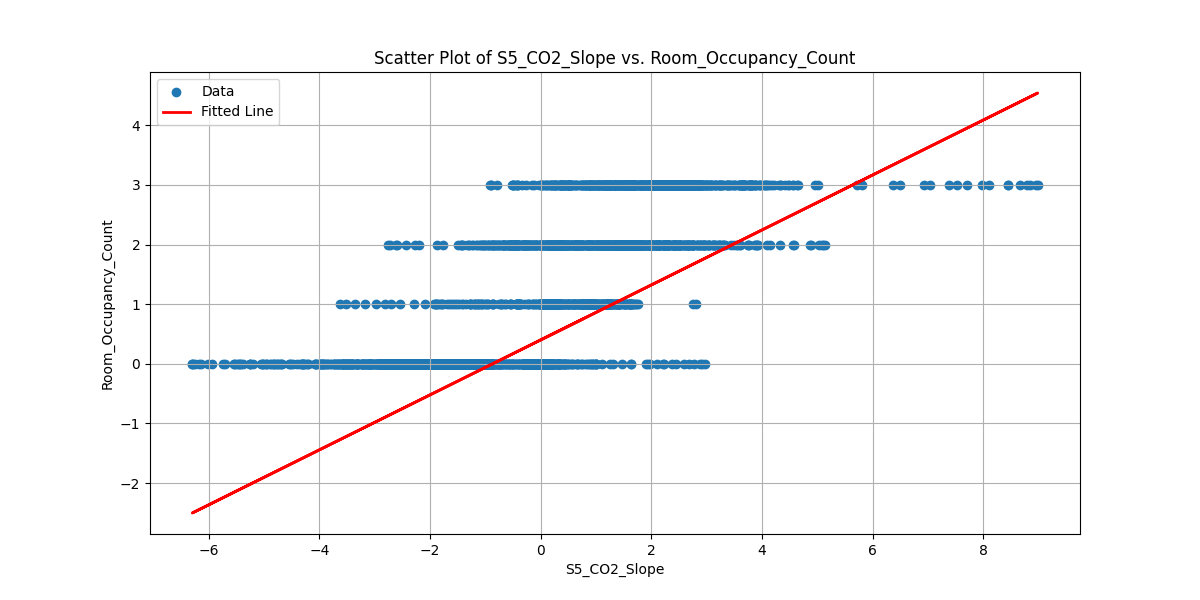}
\caption{Scatter Plot and Linear Fit of S5\_CO2\_Slope}\label{fig:S5_CO2_Slope_linear}
\end{figure}

In this figure \ref{fig:S5_CO2_Slope_linear}, we present a scatter plot depicting the relationship between the `S5\_CO2\_Slope' and the `Room\_Occupancy\_Count'. Each data point on the plot represents a specific measurement, where the x-axis corresponds to the `S5\_CO2\_Slope' values, and the y-axis represents the `Room\_Occupancy\_Count'.

Upon visual inspection, it becomes evident that there is a discernible linear increasing trend in the data points. This trend indicates that as the `S5\_CO2\_Slope' value increases, there is a corresponding rise in the `Room\_Occupancy\_Count'. The linear regression line fitted to the data further reinforces this observation.

The linear regression analysis provides numerical support for the observed trend. The intercept value of approximately 0.4008 indicates the expected `Room\_Occupancy\_Count' when the `S5\_CO2\_Slope' is zero. The positive slope of approximately 0.4611 quantifies the rate at which the `Room\_Occupancy\_Count' increases concerning changes in the `S5\_CO2\_Slope' This suggests that, on average, for each unit increase in the `S5\_CO2\_Slope' we can anticipate an increase of approximately 0.4611 in the `Room\_Occupancy\_Count'.

This observation of a linear association between the `S5\_CO2\_Slope' and the `Room\_Occupancy\_Count' aligns with the idea that the dataset may exhibit linear spatial and temporal patterns. These linear patterns may be indicative of straightforward cause-and-effect relationships within the data. In cases like these, traditional machine learning models, such as linear regression, can effectively capture these patterns without the need for specialized spatial or temporal modeling techniques.

It's important to note that while we can observe these linear patterns, a comprehensive understanding of the data collection, recording, and sensor aggregation procedures is essential for precise interpretation. Without such knowledge, we may not be able to fully elucidate the underlying mechanisms driving these patterns.

In summary, the robust performance of traditional machine learning models, even in the absence of explicit consideration for the temporal dimension, can be attributed to the presence of feature redundancy, the simplicity of linear spatial and temporal patterns, and the advantages offered by a high-frequency data sampling approach. While these findings are intriguing, it is essential to remain open to the possibility that explicitly modeling the temporal dimension could unveil deeper insights or enhance predictive capabilities in certain scenarios.

\subsection{Conclusion and Future Work}
In our investigation of room occupancy prediction, we generalized the framework presented in \cite{li2023beyond}, utilizing various machine learning approaches. Differing from its application in \cite{li2023beyond}, our focus was on room occupancy prediction as a classification problem. Our study demonstrates that this framework is adaptable to classification tasks and yields excellent predictive performance. Notably, among all the methodologies we examined, the random forest model stood out with the best performance.

An interesting observation is that, when compared to the original paper of the data source \cite{singh2018machine}, our predictive performance surpassed the original paper's results, even without accounting for time dependencies. Our analysis unveiled valuable insights into the temporal aspects of our dataset, highlighting the potential redundancies and significant correlations among certain features. Particularly, the autocorrelation results from our numeric columns provided evidence of underlying temporal dynamics. Furthermore, our dataset boasted a notably high temporal sampling frequency, with data points recorded at 30-second intervals. This high-frequency data collection potentially empowers traditional models to implicitly integrate temporal and spatial information, making the most of the fine granularity of the data points.

Our findings challenge the assumption made in the original paper that incorporating time-dependent and spatial features would invariably enhance predictive model accuracy.

Conversely, we observed limitations in the data collection process outlined in the original paper \cite{singh2018machine}. In that paper, sensor nodes were deployed at corners, the middle, and the ceiling of the room to predict the number of people in the room based on sensor readings. However, the position of individuals within the room was not considered. A straightforward implication is that by taking into account the positions of people, more accurate sensor data could be obtained, potentially leading to improved prediction performance. For future research, we plan to redesign the data collection process, considering the positions of people. In this context, we may be able to predict not only the presence but also the specific number of individuals in the room, transforming the task from a simple classification problem to a more detailed occupancy estimation.

\bibliography{ref}

\bmhead{Acknowledgments}

The source code for this project is available on GitHub at the following URL: https://github.com/YinpuLi/room-occupancy-prediction. The repository contains the implementation of the algorithms and methods comparison discussed in this paper.

\end{document}